\documentclass[a4paper]{llncs}

\usepackage[bookmarks=false]{hyperref}

\usepackage[defblank]{paralist}
\usepackage{graphicx}
\usepackage{framed}
\usepackage{color}

\usepackage{amsmath}
\usepackage{amssymb}

\newcommand{\CL}{\ensuremath{\mathcal{CL}}}
\newcommand{\codiag}{\textit{C-O Diagram}}
\newcommand{\codiags}{\textit{C-O Diagrams}}

\newcommand{\conpar}{\textsf{ConPar}}
\newcommand{\ConPar}{\conpar}
\newcommand{\UPPAAL}{\textsc{Uppaal}}
\newcommand{\CV}{\textit{Contract Verifier}}

\newcommand{\TCTLalways}{\ensuremath{\forall\,\square\,}} 

\begin{document}

\title{A Web-Based Tool for Analysing\\Normative Documents in English}
\author{John J. Camilleri \and Mohammad Reza Haghshenas \and Gerardo Schneider}
\institute{Department of Computer Science and Engineering,\\
Chalmers University of Technology and University of Gothenburg, Sweden\\
\email{john.j.camilleri@cse.gu.se},
\email{mrhaghshenas@gmail.com},
\email{gerardo@cse.gu.se}
}

\maketitle

\begin{abstract}
Our goal is to use formal methods to analyse normative documents written in English, such as privacy policies and service-level agreements.
This requires the combination of a number of different elements, including
information extraction from natural language,
formal languages for model representation,
and an interface for property specification and verification.
We have worked on a collection of components for this task:
a natural language extraction tool,
a suitable formalism for representing such documents,
an interface for building models in this formalism,
and methods for answering queries asked of a given model.
In this work, each of these concerns is brought together in a web-based tool, providing a single interface for analysing normative texts in English.
Through the use of a running example, we describe each component and demonstrate the workflow established by our tool.
\end{abstract}


\section{Introduction}

Normative texts, or ``contracts'', are documents which describe the permissions, obligations, and prohibitions of different parties over a set of actions.
They also include descriptions of the penalties which must be paid when the main norms of the document are violated.
We frequently encounter such texts in the form of
privacy policies,
software licenses,
workflow descriptions,
regulations, and
service-level agreements.
Despite being written for human consumption and thus expressed in natural language,
these kinds of documents are typically long and difficult to follow, making them hard to analyse manually.
\begin{quote}
\itshape
What commitments am I agreeing to make?\\ 
Can my information be shared with third parties?\\ 
Is it possible to avoid providing credit card details?
\end{quote}
These are the kinds of questions about a contract that we may want answered, both as users and as authors.
Using text-based search to find such answers can be tedious and unreliable, for example when clauses cross-reference each other and when exceptions are present.
Our goal is to bring formal methods to this kind of natural language analysis,
packaged as a tool which is usable by non-experts
and which requires as little understanding of the underlying technologies as possible.
We do this by first modelling these documents using a suitable formalism,
which then makes them amenable to verification using standard techniques.
This includes answering queries based on a syntactic traversal of the model,
as well as using model-checking to verify temporal properties,
by converting the model to a timed automata representation.
Each of the components required for this have been described individually in previous works by the current authors~\cite{CGS16conpar,CPS14cnl,CS17jlamp}.
This paper presents a new web-based tool for the analysis of normative texts in English,
bringing each of these components together into a single interface.

The rest of the paper continues as follows.
\autoref{sec:tool-intro} introduces the structure of the tool and gives an overview of the workflow it establishes.
In \autoref{sec:example} we then present the running example which will be used throughout the paper to demonstrate the use of our tool.
\autoref{sec:front-end} describes extraction and building a model through post-editing of tabular data,
including verbalisation of an existing model using controlled natural language.
\autoref{sec:back-end} describes the analysis which can be performed on the model, by using a set of query templates which can be customised based on the current model.
The software architecture is then summarised in \autoref{sec:architecture}.
\autoref{sec:related-work} takes a look at some related work in this area, and
we conclude with a summary of the benefits and limitations of our approach in \autoref{sec:conclusion}.


\section{The \CV\ tool}
\label{sec:tool-intro}

\begin{figure}[t]
  \includegraphics[width=\textwidth]{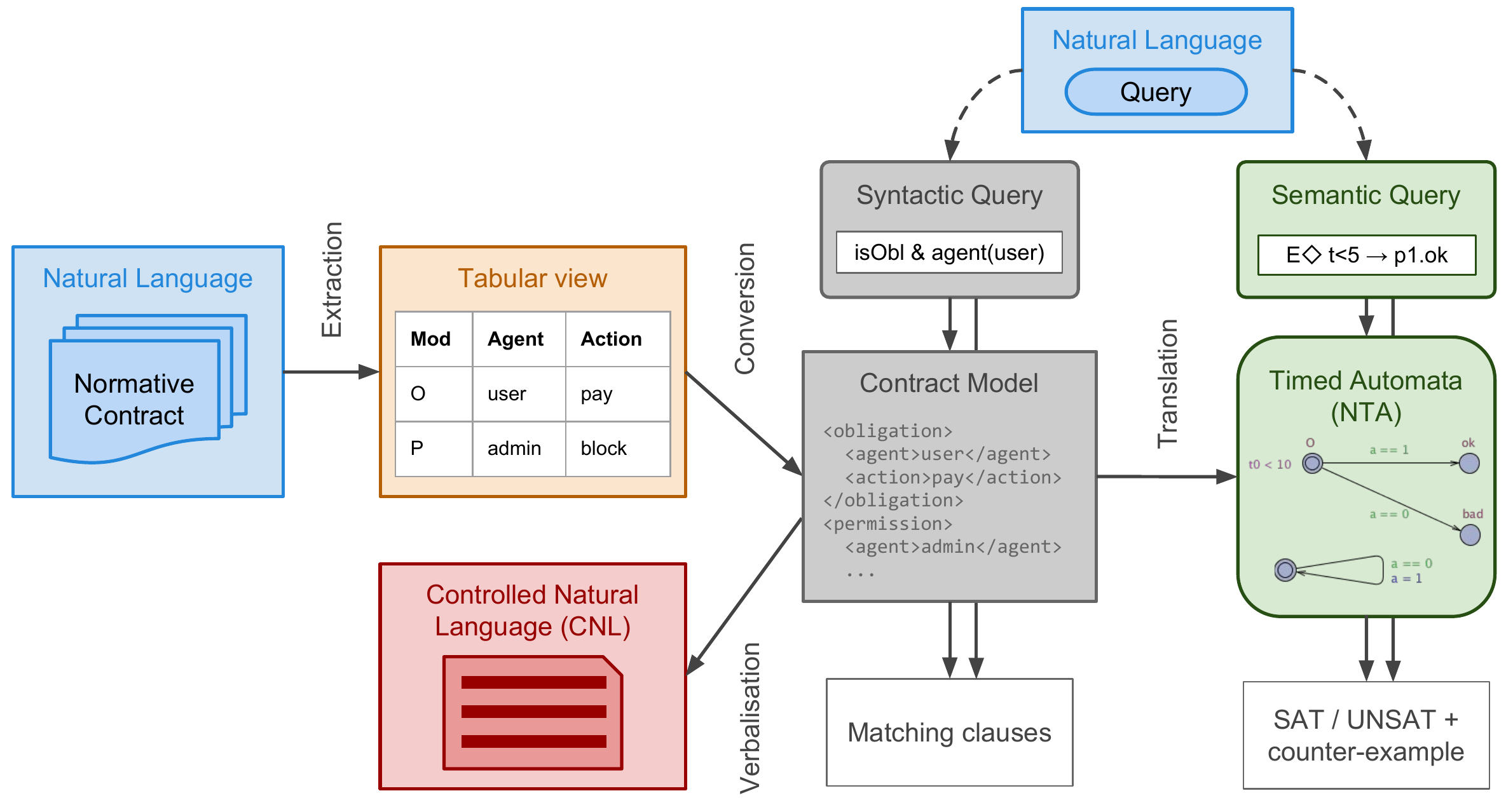}
  \caption{Overview of the contract analysis workflow established by the \CV.}
  \label{fig:system-overview}
\end{figure}

The main contribution of this work is the \CV, a web-based tool for modelling and analysing normative texts in English.
\autoref{fig:system-overview} shows an overview of the workflow which our tool covers,
summarised in the following steps:
\begin{enumerate}
  \item Users start with an English text which they wish to build a model out of.
  \item The text is submitted to an \textbf{extraction} phase which attempts to automatically extract the clauses from the text, each of which concerning at least an agent, action and modality.
  \item The results of the extraction are shown to the user in a tabular format where each cell is editable. At this point, the user must check the results of the extraction and \textbf{post-edit} the clauses which are not completely correct.
  \item Once the user is happy with the model in tabular format, this is \textbf{converted} into an actual model, which is internally represented in an XML format. 
  \item From here, the model can be verbalised in a controlled natural language and viewed in a compact formal notation.
  \item The user then performs \textbf{analysis} by selecting the queries which should be run against the model. Queries are presented as English sentence templates, which may include slots for specifying relevant arguments.
  \item The \textbf{queries} are then submitted to the server which computes the \textbf{results} and displays them to the user beside each respective query.
  \item If the answer to a query is not as expected, this could point to either:
  \begin{compactenum}
    \item a problem with the contract model, in which case the user may go back and edit the model and repeat the steps as above; or
    \item a problem with the original normative document, where the user may then modify the text and perform the analysis again, or simply leave the original formulation as it is (e.g., the analysis might detect a lack of a deadline associated with a given obligation, but this might be considered desirable by the user).
  \end{compactenum}
\end{enumerate}

A demonstration of the \CV\ is openly available on the web at \url{http://remu.grammaticalframework.org/contracts/verifier/}.
Upon visiting this URL, users will be asked to log in.
New users may create an account so that their work can be saved under their profile.
Alternatively, one may log in with a guest account, using username \verb|guest@demo| and password \verb|contract|.

\section{Running example}
\label{sec:example}

To show how our tool is used in practice, we pick a running example of a normative text
describing the rules of a university course (see \autoref{fig:case-study-text}).
This example is based on courses held at our own department,
covering the requirements for passing the course and the deadlines for the submission and grading of assignments.
It has been chosen as an example because it is concise yet contains a variety of temporal constraints and dependencies between clauses.
The text itself has been written by ourselves. 

Before publishing these rules as an official course description,
the authors (teachers/administrators) may wish to ensure that all the requirements for passing the course are enforced
and that the rules are consistent.
Once published, end users of these rules (students) may wish to query the parts which are relevant to them
or work out any flexibilities in their deadlines.
The \CV\ tool aims to meet the needs of both of these groups.

\begin{figure}[t]%
\begin{framed}%
\setdefaultleftmargin{4mm}{}{}{}{}{}%
\begin{asparablank}
\item Students need to register for the course before the registration deadline, one week after the course has started.
\item To pass the course, a student must pass both the assignment and the exam.
\item The deadline for the first assignment submission is on day 10.
\item Graders should correct an assignment within one week of it being submitted.
\item If the submission is not accepted, the student will have until the final deadline on day 25 to re-submit it.
\item The exam will be held on day 60.
\item Registered students must sign up for the exam latest 2 weeks before it is held. 
\end{asparablank}%
\end{framed}%
\vspace{-1em}%
\caption{%
Example of a normative text describing the rules involved in the running of a university course.
The course is assumed to start on day 0.
This text is used as the running example throughout this paper.%
}%
\label{fig:case-study-text}%
\end{figure}


\section{Building a contract model}
\label{sec:front-end}

\subsection{Extraction from English}
\label{sec:extraction}

The first step towards building a formal model of our natural language contract is to process the text in order to see what clause information can be extracted automatically.
We have built such a tool, named \conpar~\cite{CGS16conpar}, which can extract partial contract models from English normative texts.
\conpar{} is a natural language processing (NLP) tool, which uses the Stanford Parser~\cite{Klein03} to obtain dependency trees for each sentence in the input text.
These trees are then processed by \conpar{}, which uses the dependency representation to attempt to extract information related to:
\begin{compactenum}[(i)]
\item subject (agent)
\item verb and object (action)
\item modality (obligation, permission, prohibition)
\item temporal and non-temporal conditions.
\end{compactenum}
In addition, \conpar{} will also try to identify refinement clauses, where a single input sentence may translate to multiple sub-clauses joined together using a connective such as conjunction, choice or sequence.

\begin{figure}[t!]
\includegraphics[width=\textwidth]{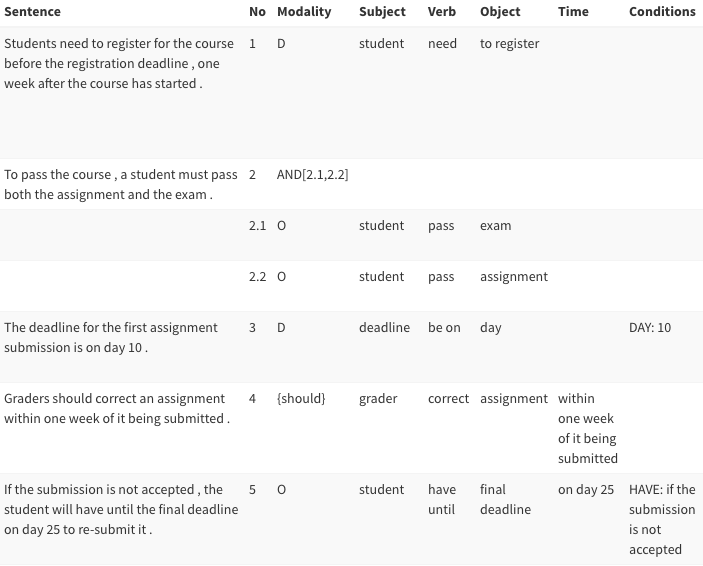}
\caption{%
Screenshot showing the output from passing our normative text through the extraction tool
(image has been cropped to improve readability).
Each row indicates a clause in the model.
Note how the second English sentence has been refined into multiple sub-clauses.
The O in the modalities column stands for \emph{obligation}, while D stands for \emph{declaration}.
}
\label{fig:conpar-output}
\end{figure}

The output of the \conpar{} tool is a tabular representation of the extracted data, which is produced in a tab-separated value (TSV) format.
This is a simple format which can be easily loaded in any spreadsheet or table-editing software.
In this representation, each row corresponds to a clause while the columns indicate the various components, as listed above.
The result of passing our example through this tool is shown in \autoref{fig:conpar-output}.

While quite a lot of clause information has been correctly extracted by the tool, there are still some errors which need to corrected manually by the user.
The use of a tabular format for displaying the output of the extraction phase facilitates this post-editing.
All cells in the table are editable, and rows can be added and deleted as needed.
While a little knowledge of this tabular format is required by the user, we nonetheless believe it is an effective way of preparing the extracted information before the model itself is created in the next step.

\subsection{Conversion to contract model}
\label{sec:conversion}

After post-editing the extracted clause information, this can be converted into a formal contract model.
The formalism used for representing contracts is based on the deontic modalities of \textbf{obligation}, \textbf{permission} and \textbf{prohibition} of agents over actions.
It includes constructs for refining clauses by conjunction, choice and sequence, and includes the possibility to specify reparations for when a clause is violated.
Clauses can be constrained by temporal restrictions or guarded based on the status of other clauses.
This formalism, which is based on \codiags~\cite{MCD+10mvs}, is described fully in~\cite{CS17jlamp}.

\begin{figure}[t!]
\includegraphics[width=\textwidth]{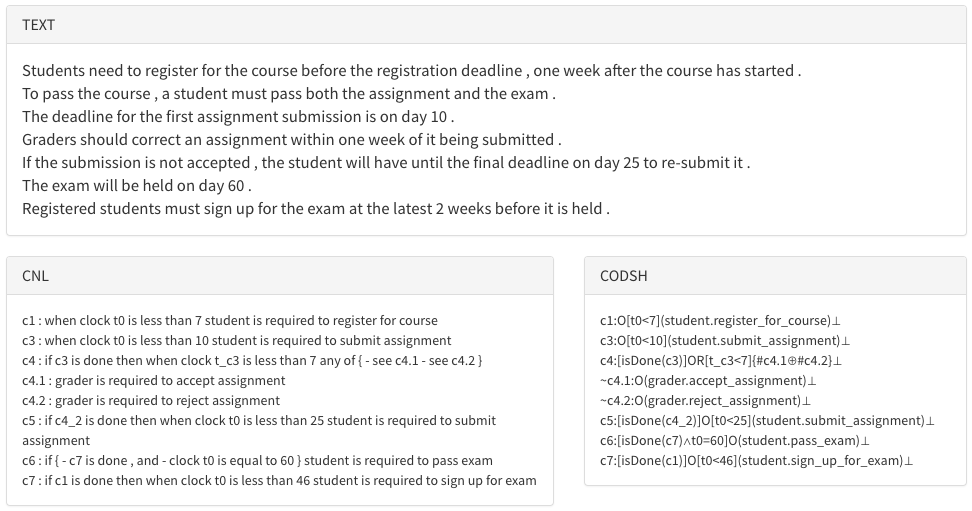}
\caption{%
Screenshot showing the different output representations of the contract model
(image has been cropped to improve readability).
Top: the post-edited input text;
bottom-left: controlled natural language;
bottom-right: compact formal notation.}
\label{fig:model-representations}
\end{figure}

This conversion step is implemented as a straight-forward script which takes the TSV representation as input and produces a contract model file.
We refer to the format of this file as COML, which is an XML-based format for storing contract models in our formalism.
Once the conversion is complete, the COML file is stored on the server and the user can view it using three different representations simultaneously:
\begin{compactenum}[(i)]
  \item post-edited text,
  \item controlled natural language (CNL), and
  \item a compact formal notation (\codiag{} Shorthand or CODSH).
\end{compactenum}
These are shown in \autoref{fig:model-representations} and explained in the sections below.

\subsection{Post-edited text}
\label{sec:post-editing}

This text corresponds to the initial column from the tabular view in the extraction phase.
While this originally comes from the input text, the individual sentences may be edited during the post-editing phase.
This means that the text displayed here is not necessarily the same as the original input text.

\subsection{Verbalisation using CNL}
\label{sec:cnl}

Given a contract model in our formalism, we have developed a method for linearising it as a phrase in a \emph{controlled natural language} (CNL)~\cite{wyner+09cnl}.
A CNL is a reduced version of a natural language (NL) which has limited syntax and vocabulary, making it in fact a formal language and thus expressible using a grammar.
CNLs are often used as interfaces for formal languages which are human-friendly, yet still unambiguous and well-defined.

The CNL designed for our contract formalism is decribed in previous work~\cite{CPS14cnl}.
We use the Grammatical Framework (GF)~\cite{Ranta2011gf} for defining the grammar for our CNL and converting to and from our internal formal representation. 
This also includes the possibility of building a contract model directly using the CNL (rather than using the extraction step), however we do not cover this input method in the present work.
The CNL representation may resemble the original NL text in some ways, however it is characterised by less variation in the expressions used and by certain structural features such as labels before each clause.

The generation of the CNL representation requires that subjects, verbs and objects are present in the lexicon.
The lexicon used is a large-scale English dictionary containing over 64,000 entries,
but if the contract contains terms which are not present in this lexicon then the generation to CNL will fail.
This failure however will not affect the usability of the rest of the tool.

\subsection{Compact formal notation}
\label{sec:codsh}

In addition to the post-edited text and the CNL, we also display a view of the model in a formal syntax.
We refer to this notation as \codiag{} Shorthand (CODSH).
It is designed mostly for developers who understand the formal structure of the contract model but would like a more condensed representation than the COML format.
This can be helpful when debugging.
In particular, this notation reveals the names which are automatically assigned to each clause in the conversion phase.


\section{Analysis}
\label{sec:back-end}

\begin{figure}[t]
\centering
\includegraphics[width=0.8\textwidth]{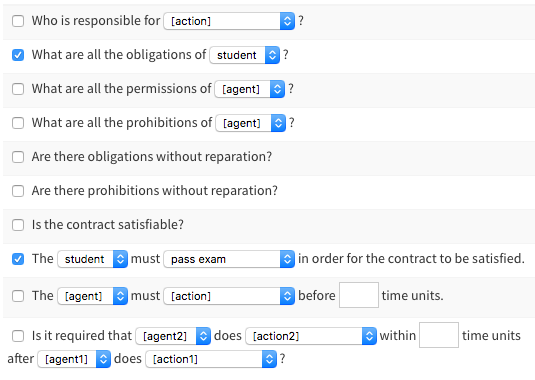}
\caption{Available query templates which users may execute against the contract model,
with drop-downs for specifying agents and actions extracted from the model.
Queries 1--6 are \emph{syntactic}, while
queries 7--10 are \emph{semantic}.}
\label{fig:queries}
\end{figure}

Once the model has been built, we can perform analysis on the contract by running queries against it.
The user is presented with a list of query templates, as shown in \autoref{fig:queries}.
Each query may have slots for parameters which the user should provide; these are either names of clauses, agents or actions.
The possible completions for these slots are extracted automatically from the contract model.

Internally, queries provided by our tool are computed in one of two ways,
either through syntactic filtering or via conversion to timed automata and using the \UPPAAL{} verification tool.
Both of these techniques are described below.

\subsection{Syntactic}
\label{sec:syntactic}

Our tool currently provides six different syntactic queries
(items 1--6 in \autoref{fig:queries}).
These queries are \emph{syntactic} in the sense that each can be solved by traversing the contract model and filtering out the corresponding clauses which match the query.
As an example, consider the following query:
\begin{quote}\itshape
What are the obligations of \emph{[agent]}?
\end{quote}
This query is internally encoded as the following conjunction of predicates:
\begin{quote}
$\mathit{isObl} \land \mathit{agentOf}([agent])$
\end{quote}
where $[agent]$ is replaced with a concrete agent name chosen by the user.
The solution to this query is computed via a Haskell function which takes the contract model as a Haskell term together with the query as a set of predicate functions, and returns a list of matching clauses.
The final output is given as a natural language sentence listing the actions corresponding to the matching clauses.
For the example, asking for all the obligations for \emph{student} will give the output below:
\begin{quote}\itshape
The following are obligations of student:
\begin{compactitem}
\item register for course
\item submit assignment
\item sign up for exam
\item pass exam
\end{compactitem}
\end{quote}
The way the result is phrased will vary based on the query, as well as the number of items in the result:
up to two results are inlined, while three or more are given as bullets.
This is to make the response more natural for the user.

\subsection{Semantic}
\label{sec:semantic}

Our tool also includes four semantic queries
(items 7--10 in \autoref{fig:queries}).
We use the term \emph{semantic} to refer to those queries which cannot be answered simply by looking at the structure of the model.
Consider the following example:
\begin{quote}\itshape
The \emph{[agent]} must \emph{[action]} before time \emph{[number]}.
\end{quote}
Determining this must take into consideration the operational behaviour of a contract model,
including when actions are performed, how new clauses are enabled and others expire.
Processing semantic queries is achieved through using model checking techniques,
by first converting a contract model into a network of timed automata~\cite{AD94tta}
and then using the \UPPAAL{} tool~\cite{LPY97uppaal} to verify temporal properties against the translated model.
This idea was introduced for \codiags{} in~\cite{DCM+14svn}; the details of our own translation can be found in~\cite{CS17jlamp}.
By using verification, we are able to quantify over all possible sequences of events with respect to the contract.

This approach requires that the query itself is encoded as a property in a temporal logic which the model checker can process.
In the case of \UPPAAL{}, the property specification language is a subset of TCTL~\cite{BDL06uppaal}.
The example query above is encoded as the following \UPPAAL{} property:
\begin{align*}
\TCTLalways &\mathit{allComplete}() \implies\\
&\big(\mathit{isDone}([agent]\_[action]) \land t_0 - Clocks\big[[agent]\_[action]\big] < [number]\big)
\end{align*}
Here, $\mathit{allComplete}$ and $\mathit{isDone}$ are helper functions included in the translated \UPPAAL{} system
which allow the status of clauses and actions to be queried from within the timed properties.
$t_0$ is a never-reset clock representing global time, while the $Clocks$ array contains a clock for each action in the system,
which is reset when that action is performed.
The expression $t_0 - Clocks\big[a\big]$ thus gives the absolute time at which action $a$ was completed.

The property is then verified using \UPPAAL, which will return a result of \textbf{satisfied} or \textbf{not satisfied}.
In cases where a symbolic trace is produced as part of the verification, this is parsed by our tool in order to provide a meaningful abstraction of it.
In this processing step, we pick out the actions performed in the trace along with their time stamps and present these as part of the result to the user.
For example, when running the query below on our contract example:
\begin{quote}\itshape
The \emph{student} must \emph{register for course} before time \emph{5}.
\end{quote}
we get the following result in the tool:
\begin{quote}\itshape
NOT Satisfied\\
The property is violated by the following action sequence:\\
- student register for course at time 6\\
- student submit assignment at time 6\\
- ...
\end{quote}
where the remainder of the trace contains the other obliged actions at time 6 or later.
This is in fact as we expect; the contract states that students have up to~7~days to register for the course,
and thus it is not true that they must have registered before day 5 in order to satisfy the entire contract.
If we change the time value in the query from 5 to 7, then the result returned is \emph{Satisfied}.




\section{Software architecture}
\label{sec:architecture}



The \CV\ tool is implemented as a PHP web application, using a MySQL database for storing user accounts and the query templates available in the system.
Contract models in COML and \UPPAAL-XML format are saved as files on the server.
No server-side framework is used.
The client-side interface is based on the AdminLTE Control Panel Template\footnote{\url{https://almsaeedstudio.com/}},
and the tabular editing interface makes use of the editableTable jQuery library\footnote{\url{http://mindmup.github.io/editable-table/}}.

The \ConPar{} extraction tool is written in Java, primarily because it uses the Stanford parser which is also implemented in Java.
The core of our system is written in Haskell,
using algebraic data types to define the structure of a contract model.
The conversion from TSV and translation to NTA are thus also written as Haskell functions to and from this data type.
The linearisation of a contract model to CNL uses the GF runtime, which is a standalone application.
Similarly, executing semantic queries requires running \UPPAAL{} as an external process.

Because of the variety of languages and programs used in our tool chain,
we provide a convenient layer over these components in the form of a small server application which provides these separate functionalities as individual web services.
This modular approach allows the web application providing the user interface to consume each component via a web API,
removing limitations on implementation language and hosting requirements, and allowing a clean separation of concerns between front-end and back-end.

\autoref{tab:web-service-api} shows a summary of the API covering all the web services provided by the server.
This API is fully documented and publicly accessible at the following URL: \url{http://remu.grammaticalframework.org:5446/}.
The server itself is also implemented in Haskell.

\begin{table}[t]
\caption{API of services provided in the web server, showing the relevant URL path and input/output formats for each service.
The various formats are: TSV (tab separated values),
COML (Contract-Oriented XML),
CODSH (\codiag\ Shorthand),
CNL (Controlled Natural Language),
and \UPPAAL-compatible XML.}
\label{tab:web-service-api}
\centering
\bgroup
\def\arraystretch{1.1}
\setlength\tabcolsep{0.35em}
\begin{tabular}{llll}
\textbf{Path}          & \textbf{Description} & \textbf{Request} & \textbf{Response} \\
\hline
\texttt{/nl/tsv}       & Clause extraction (\ConPar) & English text & TSV \\
\texttt{/tsv/coml}     & Convert TSV to COML & TSV & COML \\

\texttt{/coml/codsh}   & Show contract in shorthand & COML & CODSH \\
\texttt{/coml/cnl}     & Verbalise contract using CNL & COML & CNL \\

\texttt{/coml/syntactic} & Execute syntactic query & Query + COML & Clause names \\
\texttt{/coml/uppaal}  & Translate contract to NTA & COML & UPPAAL XML \\
\end{tabular}
\egroup
\end{table}


\section{Related work}
\label{sec:related-work}

AnaCon~\cite{ACS13fca} is a similar framework for the analysis of contracts, based on the contract logic \CL~\cite{PS07fle,PS12ddl},
which allows for the detection of contradictory clauses in normative texts using the CLAN tool~\cite{FPS09clan}.
By comparison, the underlying logical formalism we use, based on \codiags, is more expressive than \CL\ as it includes temporal constraints, cross-referencing of clauses and more.
Besides this, our translation into \UPPAAL{} allows for checking more general properties, not only normative conflicts.
In addition, their interface for specifying contracts is purely CNL-based and there is no extraction tool from English as in our case.

Information extraction from natural language is of course a field within itself,
and even in the domain of contractual documents in English there are many other works with similar goals.
The extraction task in our work can be seen as similar to that of Wyner \& Peters~\cite{WP11rer},
who present a system for identifying and extracting rules from legal texts using the Stanford parser and other NLP tools within the GATE system. 
Their approach is somewhat more general, producing as output an annotated version of the original text,
whereas we are targeting a specific, well-defined output format.
Other similar works include that of Cheng et al.~\cite{cheng+09ield},
who combine surface-level methods like tagging and named entity recognition (NER) with hand-crafted semantic analysis rules,
and Mercatali et al.~\cite{mercatali+05uml} who extract the hierarchical structure of the documents into a UML-based format using shallow syntactic chunks.

One crucial aspect in any work targeting formal analysis of natural language documents is the confidence in the extraction from a source document to the target formal language.
In our tool, the result of the extraction is usually incomplete and some amount of manual post-editing is always generally required.
Azzopardi et al.~\cite{AGP16ric} handle this incompleteness using a deontic-based logic including {\em unknowns}, representing the fact that some parts have not been fully parsed.
Furthermore, the same authors present in~\cite{AGP16inl} a tool to support legal document drafting in the form of a plugin for Microsoft Word.
Though the final objective of their work diverges from ours, 
we are both concerned with the translation of natural language documents into a formal language by using an intermediate CNL.
The main difference is that they target a more abstract formal language (very much like \CL), and as a consequence their CNL is also different.
Our formalism allows for richer representations not present in their language (e.g.~real time constraints and cross-references).
Additionally,
they do not target complex analysis of contracts (they only provide a normative conflict resolution algorithm), and
we do not provide assistance in the contract drafting process.

There is considerable work in modelling normative documents using representations other than logic-like formalisms such as our own.
%
LegalRuleML \cite{athan+15legalruleml} is a rule interchange format for the legal domain,
allowing the contents of legal texts to be structured in a machine-readable format.
The format aims to enable modelling and reasoning, allowing users to evaluate and compare legal arguments using tools customised for this format.
%
A similar project with a broader scope is MetaLex \cite{boer+08metalex}, an open XML interchange format for legal and legislative resources.
Its goal is to enable public administrations to link legal information between various levels of authority and different countries and languages, improving transparency and accessibility of legal content.
%
Semantics of Business Vocabulary and Business Rules (SBVR) \cite{SVBR1.3} uses a CNL to provide a fixed vocabulary and syntactic rules for expressing the terminology, facts, and rules of business documents of various kinds.
This allows the structure and operational controls of a business to have natural and accessible descriptions,
while still being representable in predicate logic and convertible to machine-executable form.

We note that none of the works mentioned above present a single tool for end-to-end document analysis,
starting from a natural language text and finally allowing for rich syntactic and semantic queries, as in the case of our tool.


\section{Conclusion}
\label{sec:conclusion}

In this paper we have presented \CV, a web-based tool for analysing normative documents written in English.
The tool brings together a number of different components
packaged together as a user-friendly application.
We demonstrate a typical workflow through the system, starting with an English text, extracting a contract model from it, and executing different kinds of queries against it.
Each of the components used by our tool is implemented as a standalone module, with a web-based API exposing each module as a web service.
This allows individual modules to be replaced, or new interfaces to be built, without having to make changes to the entire system.

An important feature of the \CV\ tool is the level of automation it provides: everything except the post-editing of the extracted model (and of course choosing the queries to be performed) is automatic.
For example, the names of clauses, agents and actions are automatically extracted from the contract so the user can select them using drop-down menus when making queries.
Also, each clause
is given a unique identifier as well as a {\em clock} that is reset when that clause is activated.
Though these are mainly intended for internal use when performing semantic queries,
users may even use them explicitly in the post-editing phase to encode relative timing constraints.

We see this tool as a successful implementation of a user interface for bringing together various separate components and providing a clear workflow for analysing normative texts in English.
That being said, it is as such a proof-of-concept tool and has not undergone any extensive usability testing or application in real-world scenarios.
We conclude here with a critical look at the shortcomings and limitations of the current work.

\subsection*{Evaluation}

Our goal is to make the task of analysing a normative document easier and more reliable than if one were to do it completely manually.
Measuring whether we achieve this, and to what extent, requires proper assessment.
Some evaluation has already been carried out for the natural language extraction part of the workflow (the \conpar{} tool),
measuring the accuracy of tool for extracting a correct model from a normative document.
By calculating precision and recall, $F_1$ scores of 0.49 to 0.86 were obtained for the test set of 4 documents~\cite{CGS16conpar}.

However, we currently do not have a thorough evaluation of the complete \CV\ workflow as presented here.
This would take the form of an empirical study comparing document analysis using our tool with a purely manual approach, measuring
the amount of post-editing required to build a correct model,
the time required to formulate a query and obtain a result, and
the overall ease of use of the tool in a qualitative sense.
While we consider a study of this kind important future work.

\subsection*{Limitations}


\paragraph{Extraction}
The extraction phase relies on dependency trees and thus takes a syntactic-level approach to parsing.
While a fair deal of information can be extracted in this way from simpler sentences,
a deeper understanding of a phrase often involves using related or opposite concepts which cannot be determined without more elaborate processing on the semantic level.
In addition, we assume that each input sentence translates into one or more clauses, and have no support for detecting when a phrase should actually modify an existing clause instead.

\paragraph{Modelling}
Our tool uses a tabular interface to help make the task of modelling user-friendly,
but some understanding of the underlying formal language and its semantics is necessary in order to work efficiently with it.
For example,
our formalism is essentially \emph{action-based}, where clauses prescribe what an agent should or should not \emph{do}.
However, empirically we have found that normative documents often describe what should or should not \emph{be}, i.e. referring to state-of-affairs.
While these can often be paraphrased to fit into our formalism, this is a non-trivial task which currently must be done completely by the user.

\paragraph{Verification}
When it comes to running queries, those which are syntactic can quickly be answered by an algorithm which is linear in the size of the model.
For semantic queries however, the conversion to timed automata means that the state-space explosion problem typical of model checking is a potential problem.
Certain optimisations made during the translation process could improve this somewhat.
For instance, our generated NTA contain many parallel synchronising automata as a result of the modularity of our translation,
and some {\em ad hoc} heuristics could likely be used to reduce the number of automata or the need for certain synchronisations,
thus improving the performance.
That said, this is ultimately a theoretical problem which we cannot avoid altogether.

\subsection*{Scalability}


\paragraph{Extraction}
We are limited here by the speed of the \ConPar\ tool, which itself uses the Stanford dependency parser.
While parse time is related to the length of the input sentence, in our tests based on the test data from~\cite{CGS16conpar}
we have found that parsing and extracting a single sentence takes on average roughly half a second.%
\footnote{\label{fn:test-machine} Tests carried out on a dual-core MacBook Air from 2013.}

\paragraph{Post-editing}
The tabular interface for editing the extracted clauses has no concrete limits in terms of the number of clauses it can handle.
What it does lack is support for managing a document's internal hierarchy (sections and sub-sections), which is a common feature of normative texts.


\paragraph{Analysis}
As discussed above, the running of semantic queries is the biggest barrier to the scalability of the system due to the state-space explosion problem.
For our small example here, a query requiring a search of the entire search space requires a few milliseconds to complete,
but this performance may degrade drastically as the model size increases.

\paragraph{Queries}
Our current implementation only offers a limited number of syntactic and semantic queries.
New query templates can easily be added without any theoretical constraints.
However, the user interface may need to be updated to help users navigate and possibly search through a long list of queries.
\\[12pt]
In summary, the \CV\ tool in its current state can handle documents of essentially any size in what concerns the extraction of a formal representation from a natural language text, its post-editing, and the execution of syntactic queries.
This however could be improved by adding an extra layer of document hierarchy management to the tool,
allowing the task to be segmented into smaller sub-parts.
%
In what concerns semantic queries, the tool can realistically only handle smaller individual contracts containing tens of clauses.
This is essentially due to our choice to translate contract models into \UPPAAL~timed automata.
An alternative here could be to use SAT solvers, or some other verification technology, to process the kind of semantic queries we perform.
In any case, we believe \CV\ is an important step towards a rich analysis of normative texts, even if the analysis were to be restricted to just syntactic queries.




\bibliographystyle{splncs03}
\bibliography{refs}

\end{document}